\documentclass{article}

\usepackage[final]{corl_2023} 

\usepackage[font=small]{caption}
\usepackage{csquotes}
\usepackage{amsmath}
\usepackage{graphicx}
\usepackage{float}
\usepackage{amsfonts}
\usepackage{enumitem}

\newcommand{\norm}[1]{\left\lVert#1\right\rVert}

\usepackage[dvipsnames]{xcolor}

\title{Simultaneous Learning of \\ Contact and Continuous Dynamics}

\author{
  Bibit~Bianchini, Mathew~Halm, and Michael~Posa\\
  GRASP Laboratory, University of Pennsylvania \\
  \texttt{\{bibit, mhalm, posa\}@seas.upenn.edu}
}

\begin{document}
\maketitle

\vspace{-2em}
\begin{abstract}
    Robotic manipulation can greatly benefit from the data efficiency, robustness, and predictability of model-based methods if robots can quickly generate models of novel objects they encounter.  This is especially difficult when effects like complex joint friction lack clear first-principles models and are usually ignored by physics simulators.  Further, numerically-stiff contact dynamics can make common model-building approaches struggle.  We propose a method to simultaneously learn contact and continuous dynamics of a novel, possibly multi-link object by observing its motion through contact-rich trajectories.  We formulate a system identification process with a loss that infers unmeasured contact forces, penalizing their violation of physical constraints and laws of motion given current model parameters.  Our loss is unlike prediction-based losses used in differentiable simulation.  Using a new dataset of real articulated object trajectories and an existing cube toss dataset, our method outperforms differentiable simulation and end-to-end alternatives with more data efficiency.  See our project page for code, datasets, and media:  \href{https://sites.google.com/view/continuous-contact-nets/home}{https://sites.google.com/view/continuous-contact-nets/home}
\end{abstract}



\begin{figure}[h]
    \centering
    \vspace{-0.2em}
    \includegraphics[width=0.9\linewidth]{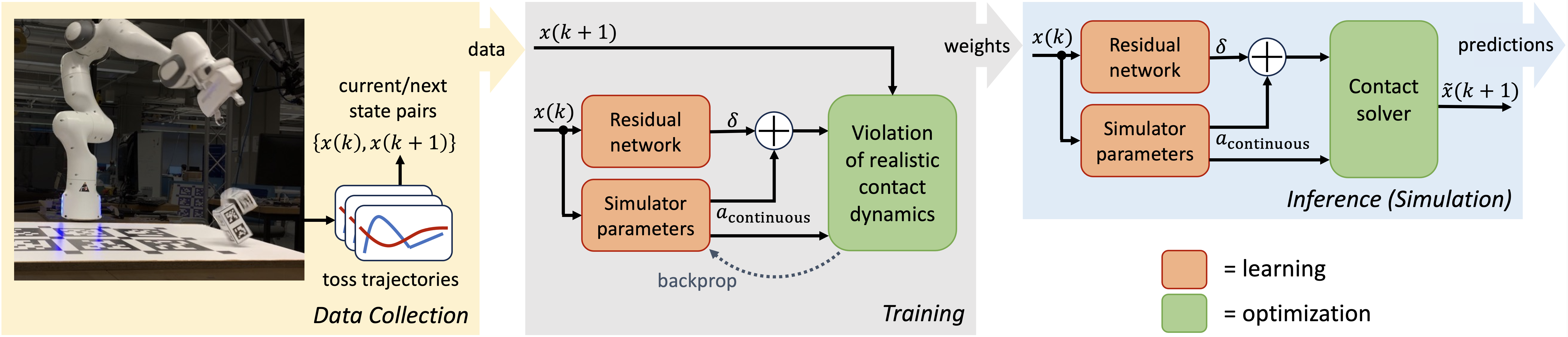}
    \vspace{-0em}
    \caption{Our method for learning dynamics of an unknown object.  \textbf{Left:} A Franka Panda automates data collection by tossing an object onto a table as the object's configuration is recorded.  \textbf{Middle:} Our violation-based implicit loss without explicit simulation trains simulator parameters and a residual $\delta$ that augments the learned continuous acceleration, encouraging the residual to learn smooth accelerations characteristic of continuous dynamics, while contact-related parameters implicitly define stiffer contact dynamics.
    \textbf{Right:} The trained model can be used with any simulator (contact solver) during inference for performing dynamics predictions.}
    \label{fig:diagram}
    \vspace{-1em}
\end{figure}

\section{Introduction}

In the future of robotic manipulation off the assembly line, robots will encounter new objects in their environment and be expected to perform useful tasks with them, such as cooking with kitchen utensils, using tools, opening doors, and packing items.  Model-based control methods work increasingly well in contact-rich scenarios \citep{aydinoglu2023consensus, aydinoglu2022real}, but rely on models of the manipulated objects.  Unlike factory settings where everything can be precisely modeled, or locomotion where the robot itself is typically the only dynamic agent, a challenge of manipulation in the wild lies in the unknown properties of the objects to be manipulated.  Model-free methods are viable, though potentially require prohibitive amounts of data \citep{ljung2010perspectives}.  Building models on the fly could enable model-based control and result in more generalizable and robust performance, but is only realistic if model-building is fast.

Manipulation is fundamentally contact-rich, and the resulting discontinuous dynamics can make model construction particularly challenging \citep{parmar2021fundamental}.  Standard system identification methods work well for identifying smooth continuous dynamics parameters even in the presence of stiff contact \citep{de2018end, fazeli2017parameter}, though assuming contact-related parameters are known.  \citet{pfrommer2020contactnets} developed a physics-based method which circumvented numerical difficulties directly by leveraging the physical structure to derive a  smooth, violation-based loss, though assuming continuous dynamics are known.  It is the aim of this work to learn both continuous and contact dynamics simultaneously.

The challenge of jointly learning continuous and contact dynamics is that the overall dynamics inherit the stiffness of contact, whose impacts can overpower the smaller, smooth, albeit important continuous accelerations.  Thus, separating continuous dynamics from contact events, while desirable, is not straightforward.  We extend \citet{pfrommer2020contactnets} which handles the contact-related portions, then combine optimization-friendly inertial parameterizations \citep{rucker2022smooth} with common deep neural network (DNN) practices of encouraging smoothness to handle the continuous dynamics via residual physics, all while enjoying the data efficiency of an implicit model with suitable loss \citep{bianchini2022generalization}.

\subsection{Contributions and Outline}

We make the following contributions in this work:
\begin{itemize}[leftmargin=0.4cm]
    \vspace{-0.4em}
    \item Present and make available a dataset of over 500 real toss trajectories of an articulated object, whose state-dependent continuous dynamics are more complicated than single rigid bodies.
    \vspace{-0.4em}
    \item Extend prior work on learning contact dynamics \citep{pfrommer2020contactnets} by simultaneously learning continuous dynamics via a combination of model-based parameterization and DNN residual physics.
    \vspace{-0.4em}
    \item Demonstrate effective performance of our method on two real datasets and two simulation scenarios with imposed actual-to-modeled continuous dynamics gaps.  We provide comparisons with differentiable simulation and end-to-end learning as alternatives.
\end{itemize}
We ground our motivations and methods in \S \ref{sec:background} with related background.  \S \ref{sec:models} details model representations, followed by loss formulations in \S \ref{sec:loss}.  With experimental setup in \S \ref{sec:experiments}, we present and discuss the results in \S \ref{sec:results}, followed by a conclusion in \S \ref{sec:conclusion} and discussion of limitations and future work in \S \ref{sec:limitations}.

\section{Background and Related Work}
\label{sec:background}

\paragraph{Challenges of contact-rich dynamics.}  Detecting contact events is extremely difficult in many practical scenarios. Many model building works  solve simpler variations, e.g. utilizing a contact detection oracle \cite{hochlehnert2021learning}, assuming knowledge of contact distances \cite{de2018end, fazeli2017parameter, allen2022learning}, or operating on simple geometries like spheres or 2D interactions \cite{fazeli2017parameter}. Our work builds off \cite{pfrommer2020contactnets}'s contact-related parameter learning that performs automatic segmentation of contact/non-contact effects, without access to an oracle to identify contact events.  Our work extends this contact-implicit model building beyond contact dynamics, to building full dynamics models, without impractical contact detection assumptions.

\paragraph{Implicit representations for discontinuous functions.}
Recent works have employed implicit approaches to represent sharp functions smoothly, whether those functions represent discontinuous control policies \citep{florence2022implicit} or contact models \citep{fazeli2017learning}.  Other works demonstrating differentiation through these implicit representations make them viable for use in learning \citep{agrawal2019differentiating, amos2017optnet}.  These techniques rely on smooth parameters to implicitly encode signals for discontinuous or extremely stiff events, which accurately characterizes contact dynamics.  However, the implicit model representation can be data inefficient to optimize when combined with explicit losses \citep{bianchini2022generalization}.  Implicit model representations in combination with an informative loss can successfully learn contact parameters \citep{pfrommer2020contactnets}.

\paragraph{Differentiable simulation.}
Widely used for policy learning and control, differentiable simulators are also useful for system identification \citep{de2018end, le2023single, howell2022dojo}.  Differentiable simulators use governing dynamics that can be explicitly differentiated, and often compare simulated predictions with observed motion, using the difference to supervise model training.  However because they use prediction-based losses, differentiable simulators notoriously can have difficult to optimize loss landscapes when identifying contact-related parameters \citep{antonova2023rethinking}.  Differentiable simulation can be combined with artificially soft contact models to improve optimization \citep{underactuated}, at the cost of model accuracy \citep{parmar2021fundamental}.

\paragraph{Inertial parameterizations.}
The inertia of a rigid body is completely described by 10 parameters:  the mass, center of mass (3), and moments/products of inertia (6).  Learning these directly can be problematic since many members of $\mathbb{R}^{10}$ are physically infeasible inertia vectors.  There are several previously developed mappings from $\theta_\text{inertia} \in \mathbb{R}^{10}$ to physically feasible sets of inertial parameters \citep{rucker2022smooth, sutanto2020encoding, atkeson1986estimation}, and thus learning $\theta_\text{inertia}$ to indirectly yield inertial properties becomes a well posed optimization problem.  We use the \citet{rucker2022smooth} parameterization in this work.

\paragraph{Residual physics.}
While model-based structures typically boast data efficiency compared to model-free approaches \citep{ljung2010perspectives}, they fundamentally suffer from inaccuracies of the model on which they are based.  Residual physics \citep{wong2022oscar, heiden2021neuralsim, ng_2006, ajay2018augmenting, zeng2020tossingbot} mitigates this by learning an expressive residual that fills a data efficient but possibly insufficient structured model's sim-to-real gap.  We use a residual physics DNN in this work to specifically augment the continuous dynamics of our structured model.


\section{Model Representations}
\label{sec:models}

We consider a discrete dynamics model $f$ parameterized by a set of learnable parameters $\theta$ that takes in some state $x(k)$ and set of control inputs $u(k)$ and performs a single simulation step,
\begin{align}
	x(k+1) &= f^\theta (x(k), u(k)).
\end{align}
This makes no assumptions about the structure of $f$ or what the learned parameters $\theta$ represent.  In an unstructured case, $f$ could be learned as a DNN where $\theta$ is the weights and biases of the network.  In a more structured case (e.g. rigid body dynamics), $\theta$ represents physical parameters of the system.  Numerical methods commonly simulate contact by introducing an optimization problem to search for contact impulses $\lambda(k)$ from a feasible set of contact impulses $\Lambda$ over the time step,
\begin{subequations} \label{eqn:implicit_inference}
\begin{align}
    x(k+1) &= g^\theta(x(k), u(k), \lambda(k)), \\
    \text{where} \qquad \lambda(k) &= \arg \min_{\lambda \in \Lambda} \, h^\theta \left( x(k), u(k), x(k+1), \lambda \right),
\end{align}
\end{subequations}
where $h^\theta$ measures violation of contact constraints.  
This generic formulation underpins many common simulators, where the embedded optimization problem may be a linear complementarity problem (LCP) \citep{anitescu1997formulating, stewart1996implicit}, a second-order cone program \citep{castro2022unconstrained}, or some more generic structure \citep{todorov2012mujoco}.

\subsection{Measuring Violation of Rigid Body Contact Dynamics}
Inspired by the LCP formulation from \citet{stewart1996implicit}, we follow standard methods for conversion to an equivalent optimization problem form in \eqref{eqn:implicit_inference}, introduced by \citet{pfrommer2020contactnets}.  First, we let $\Lambda$ describe a Coulomb friction cone,
\begin{align}
    \lambda \in \Lambda \quad \Leftrightarrow \quad \norm{\lambda_{t, i}} \leq \mu_i \lambda_{n, i} \quad \forall i = 1, \dots p, \label{eqn:lambda_constraint}
\end{align}
for a system with $p$ contacts.  Then we use penalty terms to describe violation of force complementarity, energy dissipation, and geometric penetration for each contact $i$,
\begin{subequations}
\begin{align}
    h_{\text{comp}, i}^\theta (k) &= \lambda_{n, i}(k) \phi_i(k+1), \label{eqn:comp} \\
    h_{\text{diss}, i}^\theta(k) &= \lambda_i^T(k) \begin{bmatrix} \mu_i \norm{ J_{t,i}(k) v(k+1)} \\ J_{t,i}(k) v(k+1) \end{bmatrix}, \label{eqn:diss} \\
    h_{\text{pen}, i}^\theta(k) &= \min \left( 0, \phi_i(k+1) \right)^2, \label{eqn:pen}
\end{align}
\end{subequations}
where $\phi$ is the set of signed distances, $\mu$ is set of friction coefficients, $J = [J_n; J_t]$ is the normal and tangential contact Jacobians, and $x = [q; v]$ is the state of system configuration and velocity.  Thus, with relative weighting between the terms, $h^\theta$ becomes
\begin{align}
    h^\theta(k) &= \sum_{i=1}^p \sum_{j \in \{ \text{comp}, \text{diss}, \text{pen} \}} w_j h^\theta_{j, i}(k). \label{eqn:h_form}
\end{align}
With the contact dynamics described by $h$ in \eqref{eqn:h_form} and the constraint in \eqref{eqn:lambda_constraint}, the function $g$ is a discretized version of Newton's third law to update system velocities, and an implicit Euler step to update the configuration.  This $g$ is a function of the implicit variable $\lambda$, and can be written as
\begin{subequations} \label{eqn:pred}
\begin{align}
    v(k+1) &= v(k) + a_\text{continuous} \Delta t + M^{-1} J^T \lambda(k), \\
    q(k+1) &= q(k) + \Gamma v(k+1) \Delta t, \label{eqn:config_pred}
\end{align}
\end{subequations}
where $a_\text{continuous}$ is the acceleration of the system due to continuous dynamics, and $\Gamma$ maps velocity-space to configuration-space (e.g. mapping angular velocity to the time-derivative of an orientation quaternion).  See Appendix \ref{apx:model_tuning} for the selection of the introduced hyperparameter weights in $h^\theta$.

\subsection{Learnable Parameters}
\label{subsec:parameters}

With the above model structure, the learnable parameters include the following. \textbf{Geometry} determines $J$ and $\phi$, both functions of the system configuration, i.e. $J(q(k)), \phi(q(k))$.  We parameterize $J$ and $\phi$ by a set of vertices whose 3D locations are learnable.  \textbf{Friction}, via $\mu$, determines the permissible set of contact impulses per contact point.  In this work, the friction is parameterized by a single scalar $\mu$.  \textbf{Inertia} affects the model's forward predictions in \eqref{eqn:pred}, where $M$ and $a_\text{continuous}$ appear.  With articulation, $M$ is a function of the system configuration, i.e. $M(q(k))$, and $a_\text{continuous}$ of the full state, $a_\text{continuous}(x(k))$.  We map learnable parameters in $\mathbb{R}^{10}$ to a physically feasible set of 10 inertia parameters, per \citet{rucker2022smooth}.  Under autonomous dynamics, the mass of a system is unobservable if contact forces are not measured \citep{fazeli2017parameter}.  Thus, we keep the total system mass fixed, then learn the remaining moments and products proportionally as well as the center of mass.

\subsection{Residual Network}
\label{subsec:residual}

In this work, we use a residual physics DNN to compensate for inaccuracies in the model structure in \eqref{eqn:implicit_inference}.  Since rigid body contact solvers like \cite{stewart1996implicit} work reasonably well to capture real inelastic contact dynamics \citep{acosta2022validating}, we encourage the residual to fill gaps in the continuous dynamics.  We add components to the continuous acceleration of the system,
\begin{align}
    a_{\text{continuous}}\left(x(k) \right) &= a_\text{continuous, model}\left(x(k) \right) + \delta^\theta\left(x(k) \right),
\end{align}
where $\delta^\theta \in \mathbb{R}^{n_\text{vel}}$ is the output of a residual network whose input is the state of the system.  See Appendix \ref{apx:residual} for network architecture details.

Adding costs on the norm of the network's output and on its weights encourages the residual to be small and smooth, respectively, as continuous dynamics are in comparison to contact dynamics.  For end-to-end alternatives which aim to capture both continuous and contact dynamics in one network, weight regularization is no longer beneficial since it leads to unrealistically soft contact dynamics.  To capture the stiffness of contact, the result is an end-to-end network with extreme input sensitivity.  In contrast, incorporating our residual into the continuous acceleration allows the inherent stiffness of contact to be implicitly learned via geometric and frictional properties, leaving the residual in a smooth domain better suited for standard DNN approaches \citep{parmar2021fundamental}.



\section{Loss Formulation}
\label{sec:loss}

\begin{figure}[t]
    \centering
    \includegraphics[width=0.9\linewidth,trim={00mm 3mm 0mm 3.5mm},clip]{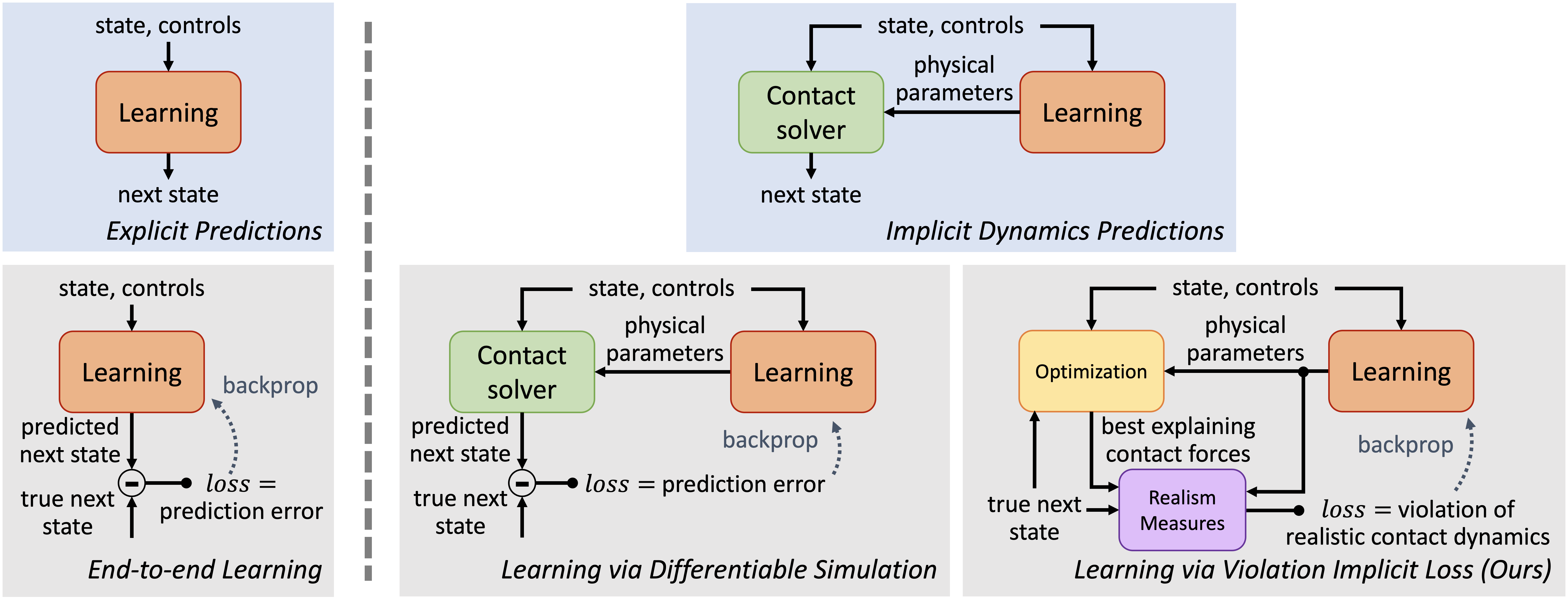}
    \caption{\textbf{Top row:} Using an explicit model (left) versus an implicit model (right) for performing dynamics predictions.  Explicit models aim to predict next states directly from current states and inputs.  Implicit models instead parameterize and leverage contact solvers, which produce next states as a result of an optimization problem.  \textbf{Bottom row:}  A common way to train an explicit model is via a prediction-based loss (left).  Implicit models can also be trained with a prediction-based loss (middle), requiring performing and differentiating through the contact solver.  Our approach trains an implicit model with a violation-based loss (right), avoiding simulation during training time and producing smoother, more informative gradients.}
    \label{fig:loss_variations}
    \vspace{-1em}
\end{figure}

Our specific model structure alone does not affect the generalization capabilities of a model, and the choice of loss function is also of vital importance for stiff function classes \citep{bianchini2022generalization}.  Fig.\ \ref{fig:loss_variations} diagrammatically illustrates the differences between the losses presented in this section and how they relate to explicit versus implicit model usage for simulation.

\subsection{Prediction Loss}
\label{subsec:pred_loss}
Standard approaches in model-building or system identification \citep{nagabandi2020deep, howell2022dojo, heiden2021neuralsim} use a prediction-based loss that penalizes the error in a candidate model's predictions,
\begin{align}
    \mathcal{L}_\text{prediction} &= \norm{x(k+1) - f^\theta \left( x(k), u(k) \right)}^2.  \label{eqn:prediction_loss}
\end{align}
Differentiable simulators typically employ the implicit optimization problem as in \eqref{eqn:implicit_inference} to solve for contact impulses, so the loss becomes (equivalently)
\begin{subequations} \label{eqn:implicit_pred_loss}
\begin{align}
    \mathcal{L}_\text{prediction} &= \norm{x(k+1) - g^\theta \left( x(k), u(k), \lambda(k) \right)}^2, \\
    \text{such that} \qquad \lambda(k) &= \arg \min_{\lambda \in \Lambda} \, h^\theta \left( x(k), u(k), x(k+1), \lambda \right).
\end{align}
\end{subequations}
Both explicit approaches \eqref{eqn:prediction_loss} and implicit approaches \eqref{eqn:implicit_pred_loss} functionally result in simulating a candidate model and penalizing the difference between its prediction and the true dynamics observation.

\subsection{Violation-Based Implicit Loss}
\label{subsec:violation_loss}

Despite the increasing prevalence of implicit approaches, prediction-based losses inhibit the generalization benefits of implicit models \citep{bianchini2022generalization}.  A violation-based implicit loss of the form
\begin{align}
    \mathcal{L}_\text{violation} &= \min_{\lambda \in \Lambda} \left[ \norm{x(k+1) - g^\theta \left( x(k), u(k), \lambda \right)}^2 + h^\theta \left( x(k), u(k), x(k+1), \lambda \right) \right],
\end{align}
uses $h$ as a soft constraint and, as a result, boasts greater data efficiency \citep{bianchini2022generalization}.  This loss itself is an optimization problem that solves for the set of contact impulses $\lambda$ that balances 1) explaining the observed motion and 2) matching the learned contact dynamics model.  Thus minimizing this loss function through the training process is a bilevel optimization problem.  For full details, see \citep{bianchini2022generalization, pfrommer2020contactnets}.  This loss performs inference over contact mode, a key enabling technique for contact-implicit planning and control \citep{aydinoglu2023consensus, posa2014direct}.

The exact form of the prediction error term $\norm{x(k+1) - g^\theta \left( x(k), u(k), \lambda \right)}^2$ employed herein penalizes errors in velocity space, since configurations are an affine function of velocity predictions \eqref{eqn:config_pred}.  A natural way to combine mixed linear and angular terms is to convert all into energy units via
\begin{align}
    l^\theta_\text{pred, energy}(k) &= \norm{ M \Delta v(k) + J^T \lambda }_{M^{-1}}^2, \\
    \text{where} \qquad \Delta v(k) &= -v(k+1) + v(k) + a_\text{continuous}\Delta t.
\end{align}

\section{Experimental Setup}
\label{sec:experiments}

\begin{figure}[t]
    \centering
    \includegraphics[width=0.215\linewidth,trim={0mm 0mm 0mm 20mm},clip]{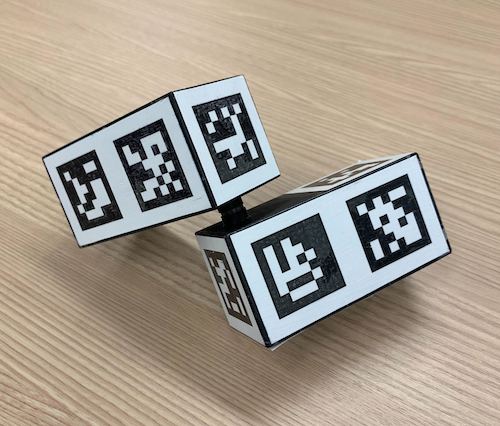}
    \includegraphics[width=0.24\linewidth, trim={10cm 0cm 30cm 109cm}, clip]{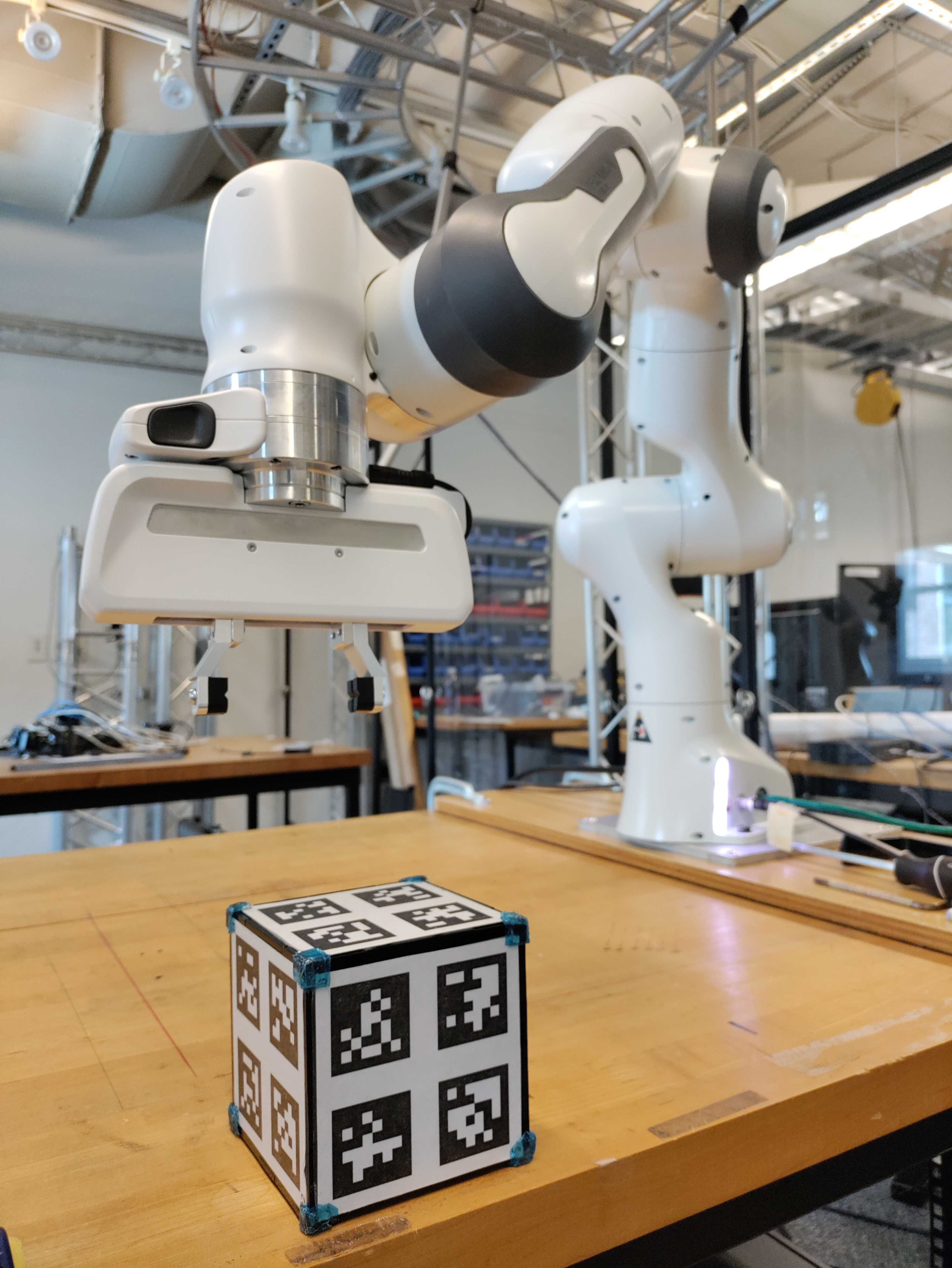}
    \includegraphics[width=0.12\linewidth,trim={0mm 0mm 0mm 0mm},clip]{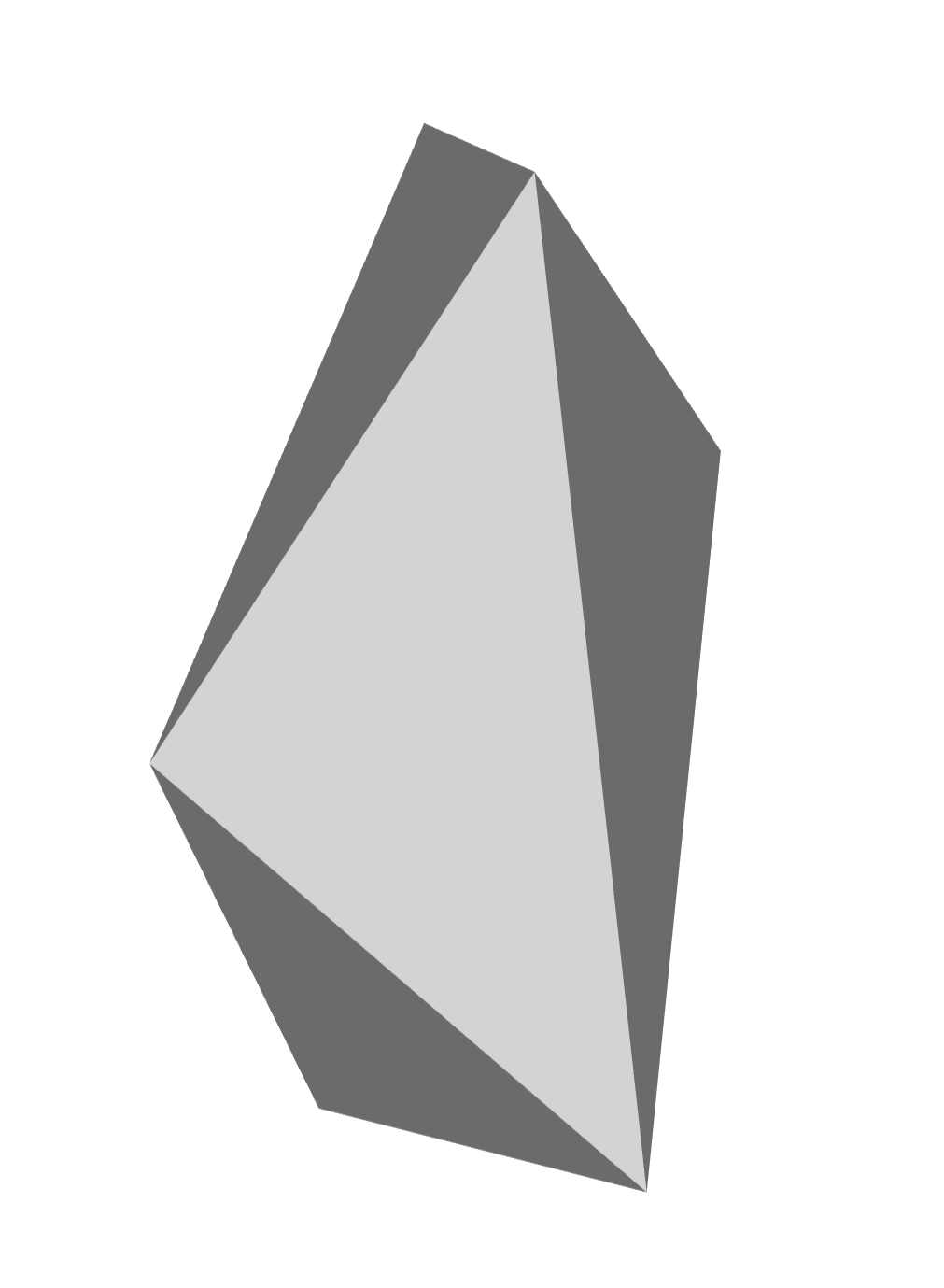}
    \includegraphics[width=0.15\linewidth,trim={0mm 0mm 0mm 0mm},clip]{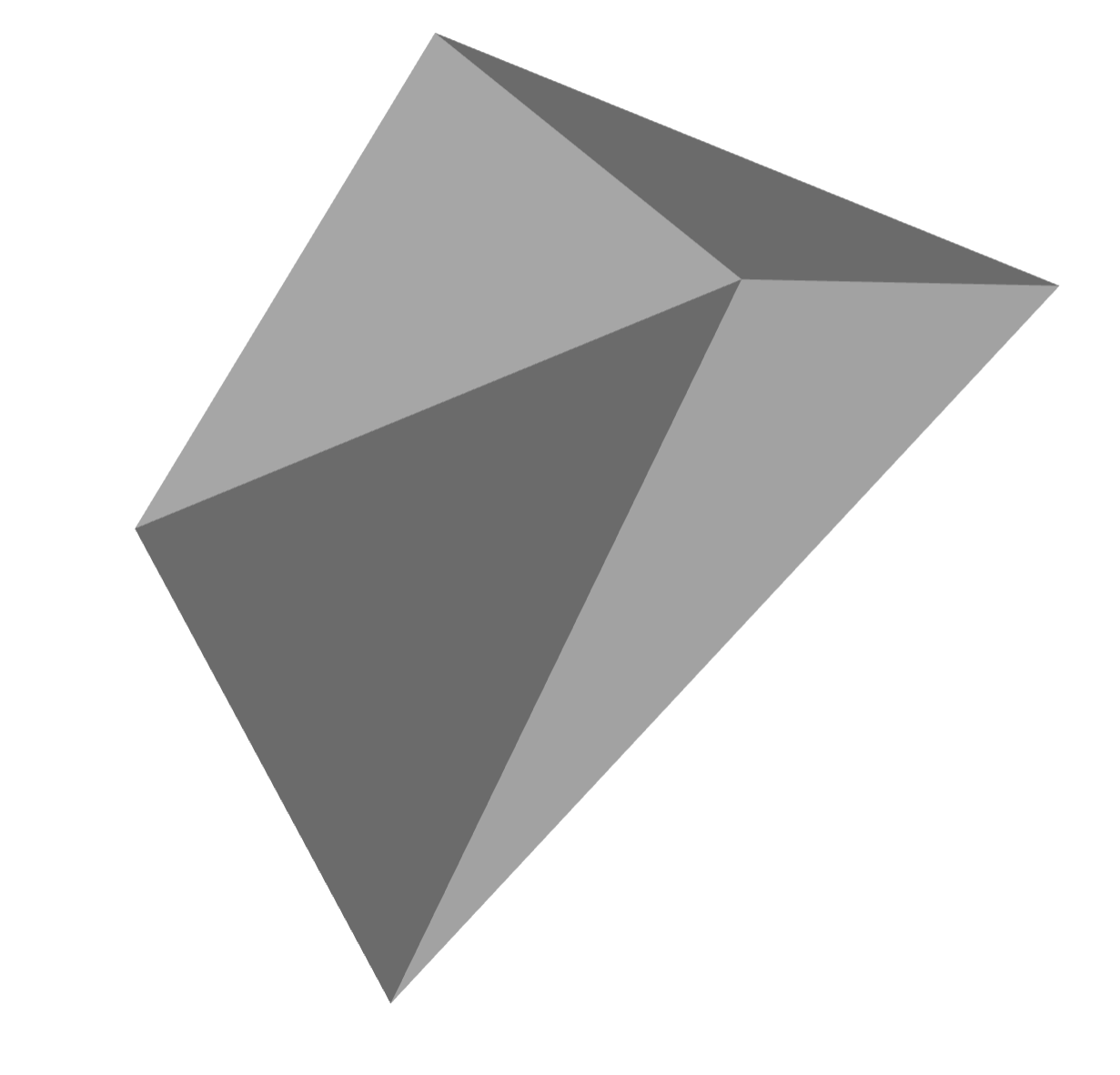}
    \vspace{-2mm}
    \caption{Our experimental systems.  \textbf{Left:} A real two-link articulated object, with each link as 5cm by 5cm by 10cm.  \textbf{Left middle:} A real cube of width 10cm, whose dataset was contributed by \citep{pfrommer2020contactnets}.  \textbf{Right:} A simulated 6-vertex asymmetric object from two views.  The volume of the asymmetric is similar to the volume of one of the articulated object links.}
    \label{fig:experiments}
    \vspace{-1.5em}
\end{figure}

For all experiments, we consider a system autonomously falling under gravity and colliding with a flat plane.  In addition to the \textbf{cube toss} dataset contributed by \citet{pfrommer2020contactnets}, we contribute one new real dataset and two simulated scenarios.  We used a Franka Emika Panda 7 degree-of-freedom robotic arm to automate the toss data collection of a \textbf{two-link articulated object}.  Pose information of each link is tracked using TagSLAM \citep{pfrommer2019tagslam}.  These two body poses are combined into minimal coordinates via an optimization problem that minimizes pose offset of both links.  While we keep the system mass fixed, our model can freely decide the mass distribution across the two links.

We add a \textbf{vortex simulation of an asymmetric object} example, simulating dynamics with a spatially-varying force field pulling towards and swirling around a fixed vertical line.  The initialized model is unaware of this continuous dynamics augmentation.  We test in this scenario with an asymmetric object with 6 vertices.  Lastly, we include a \textbf{gravity simulation of the articulated object}. This scenario features simulated dynamics of the articulated object from our new dataset with typical gravitational acceleration of $9.81 \text{m}/\text{s}^2$.  We test at a fixed training set size of 256 tosses from poor initial guesses, at some fraction $\in [0, 2]$ of this simulated gravity.  All simulation data was generated using Drake \citep{drake} and features a significant gap between the model's believed and the simulator's actual dynamics.  See Fig.\ \ref{fig:experiments} for visuals of these systems.

{\renewcommand{\arraystretch}{1.0}
\begin{table}[b]
    \centering
    \vspace{-0.8em}
    \begin{tabular}{ | c | c | c | c | }
        \hline
        \textbf{Name} & \textbf{Parameterization} & \textbf{Loss} & \textbf{Residual} \\ \hline \hline
        CCN (ours) & Structured & Violation implicit & \\
        CCN-R (ours) & Structured & Violation implicit & \checkmark \\
        DiffSim & Structured & Prediction error & \\
        DiffSim-R & Structured & Prediction error & \checkmark \\
        End-to-end & DNN & Prediction error & N/A \\ \hline
    \end{tabular}
    \vspace{0.3em}
    \caption{Tested approaches.  CCN stands for our extension of Continuous dynamics learning plus the contact dynamics learning in ContactNets \citep{pfrommer2020contactnets}.  DiffSim is Differentiable Simulation using differentiable contact dynamics defined in \citet{anitescu2006optimization}.  The -R modifier indicates residual physics is included.  DiffSim ablates our violation implicit loss function, and the End-to-end baseline ablates the physical structure imposed by the rigid body model-based parameterization.  See Appendix \ref{apx:end_to_end} for End-to-end network details.}
    \label{table:methods}
    \vspace{0em}
\end{table}
}

\begin{figure}[t]
    \centering
    \includegraphics[width=\linewidth]{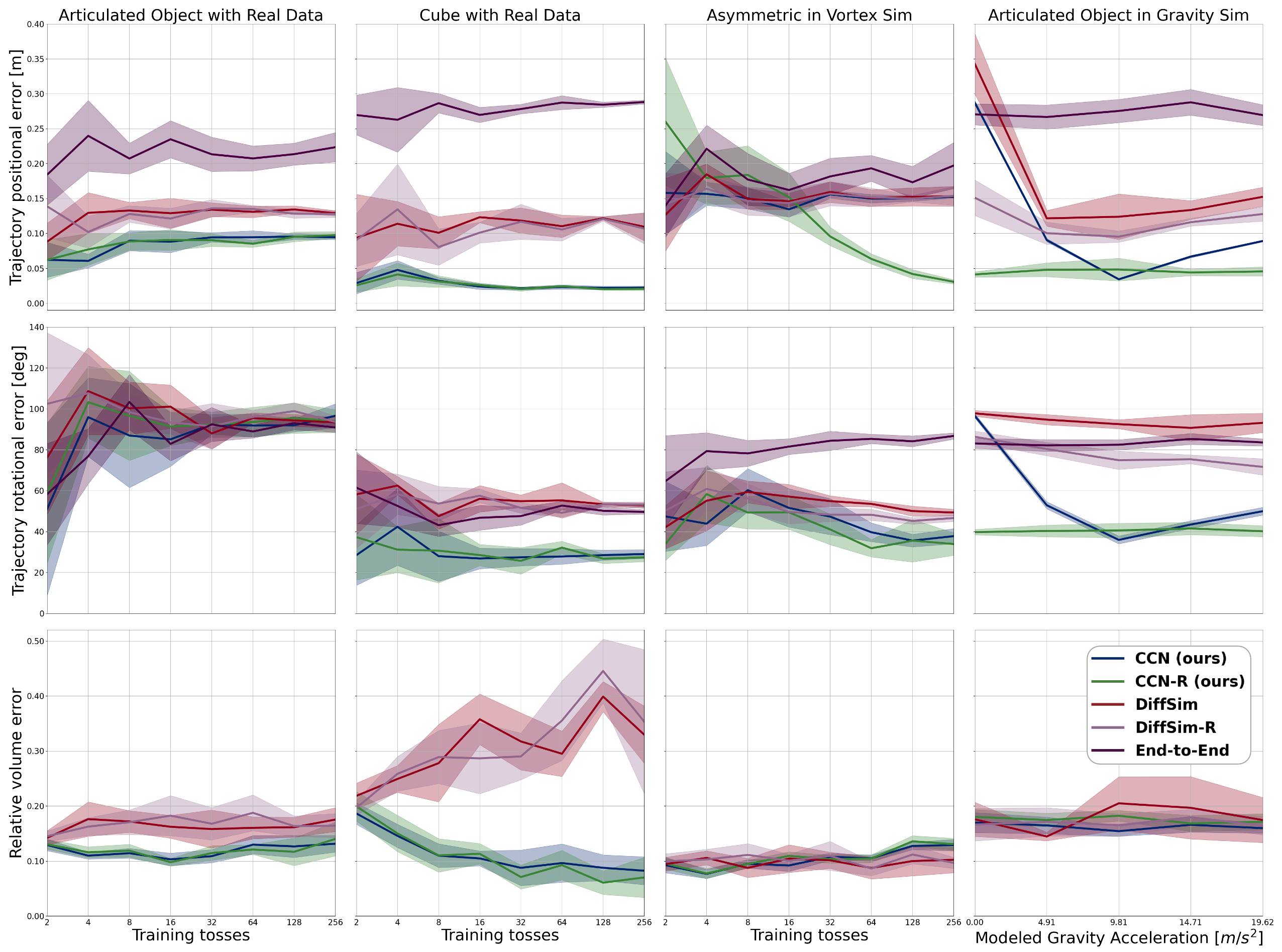}
    \caption{Results from the four experiments.  Shaded regions indicate normal t-score 95\% confidence intervals.  \textbf{Left column:} The real articulated object featured rotations that were difficult for any of the methods to capture over a long time horizon (middle), yet our CCN approaches outperforms DiffSim on geometry error (bottom) and all alternatives on positional error (top).  \textbf{Left middle column:}  For every metric on the real cube experiments, our CCN approaches outperform DiffSim and End-to-end.  \textbf{Right middle column:} While every method achieved low geometry error on the asymmetric object in simulated vortex dynamics, our CCN approaches performed the best in rotational error, and only our approach with residual (CCN-R) was able to achieve low positional error.  \textbf{Right column:} The x-axis for the gravity experiments swept over an initial modeled gravitational acceleration.  Despite poor model discrepancy, only our approach with residual CCN-R is able to maintain good performance across all metrics at different model discrepancies.}
    \label{fig:results}
\end{figure}

\begin{figure}[h]
    \centering
    \includegraphics[width=0.26\linewidth, trim={0cm 0cm 3cm 2cm}, clip]{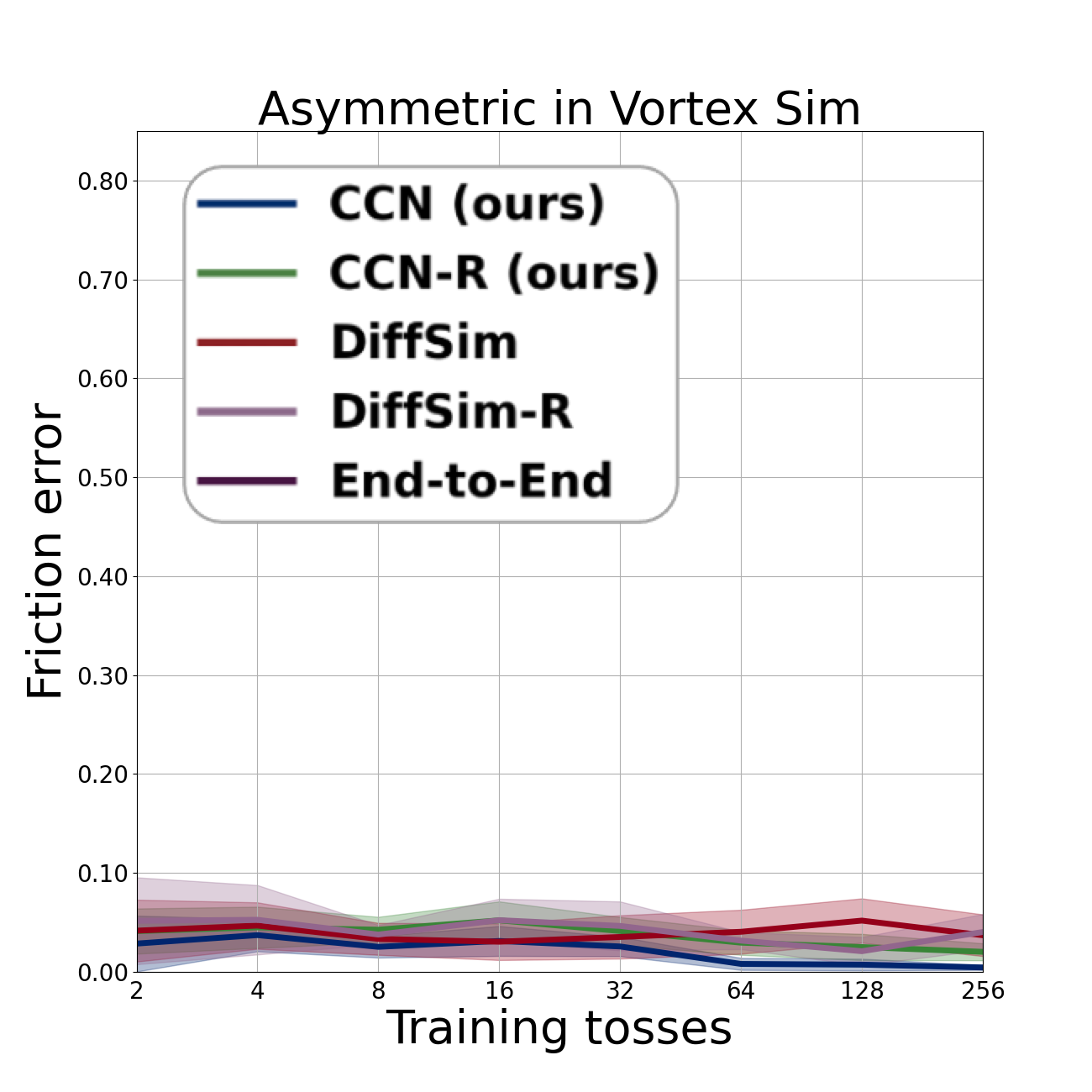}
    \includegraphics[width=0.225\linewidth, trim={4cm 0cm 3cm 2cm}, clip]{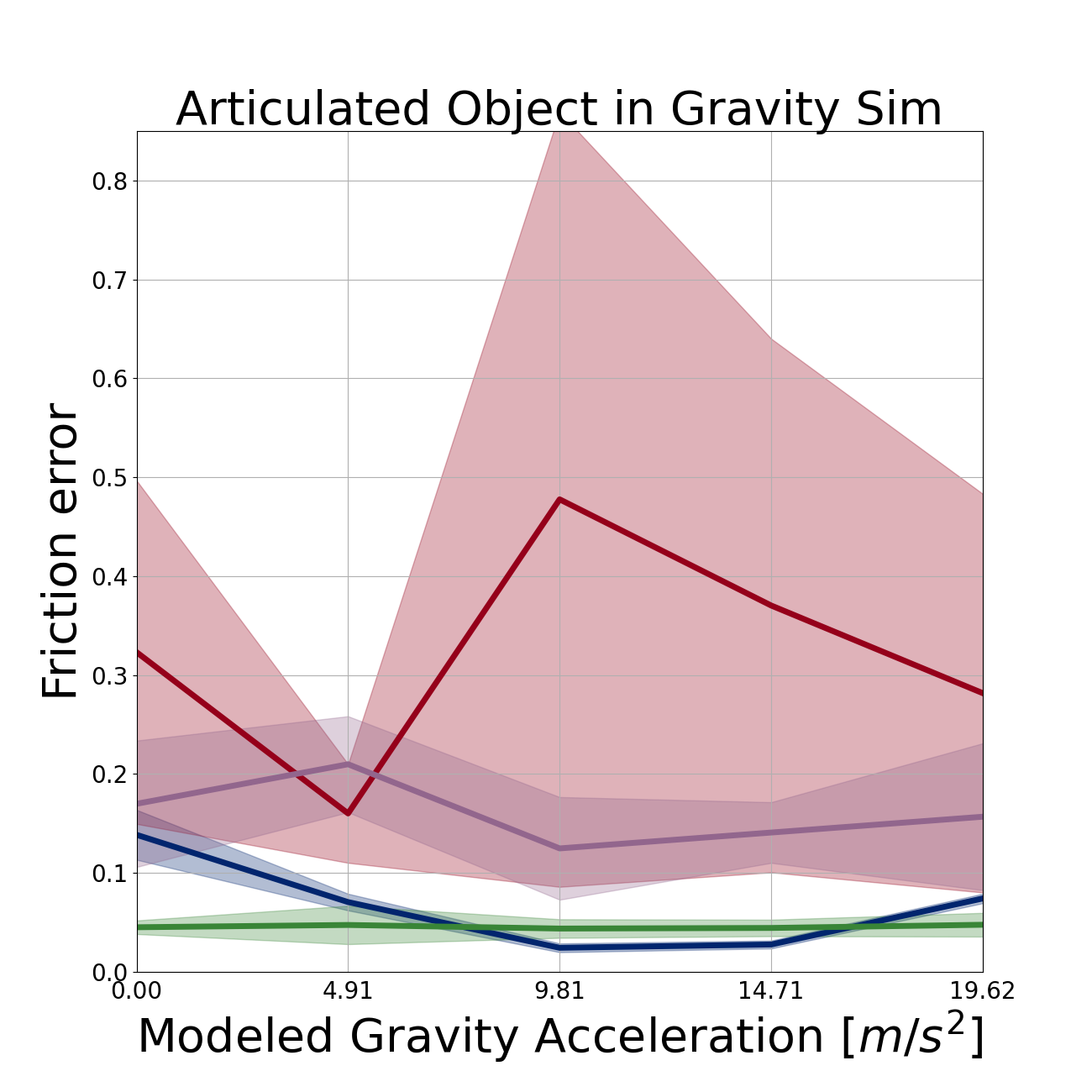}
    \,
    \includegraphics[width=0.26\linewidth, trim={0cm 0cm 3cm 2cm}, clip]{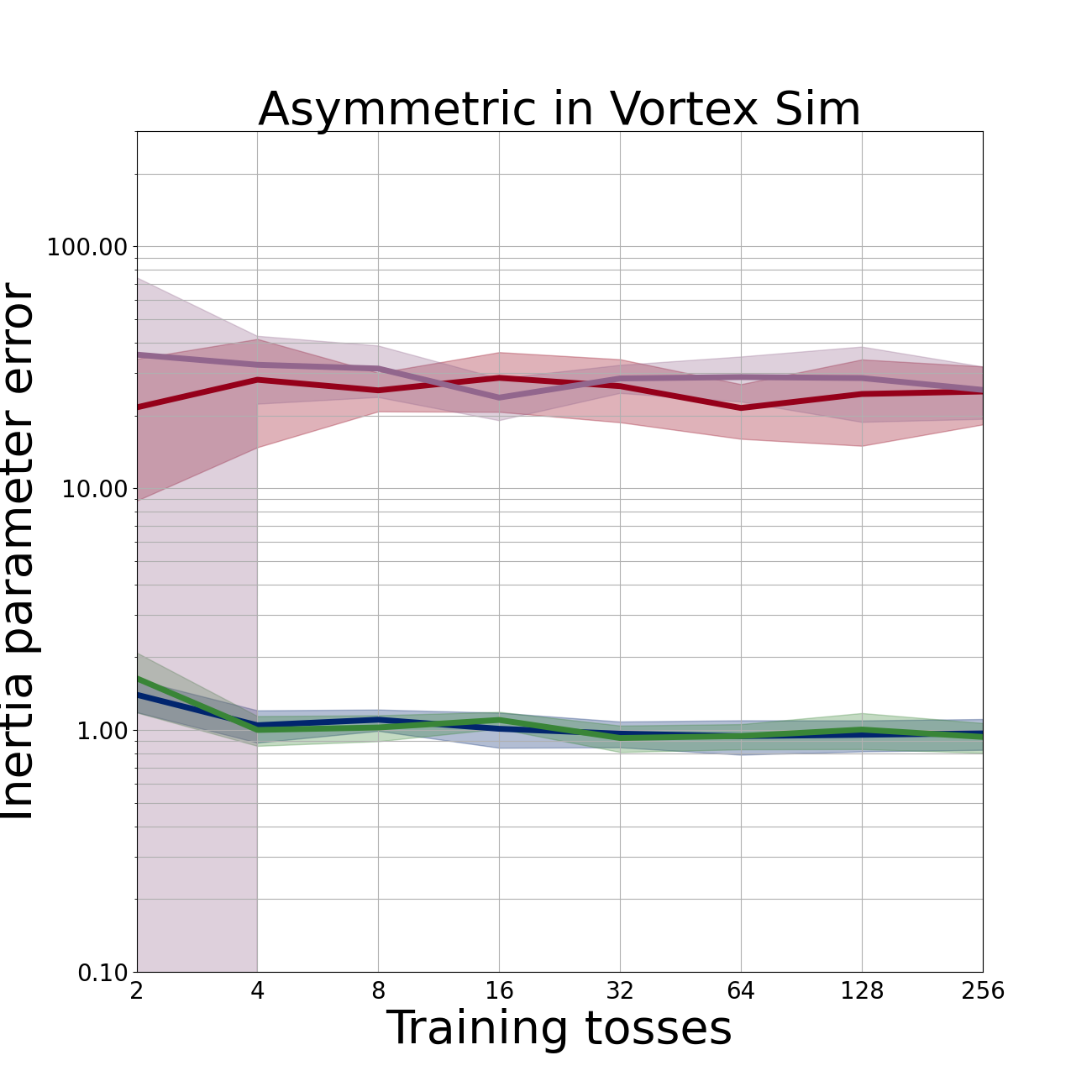}
    \includegraphics[width=0.225\linewidth, trim={4cm 0cm 3cm 2cm}, clip]{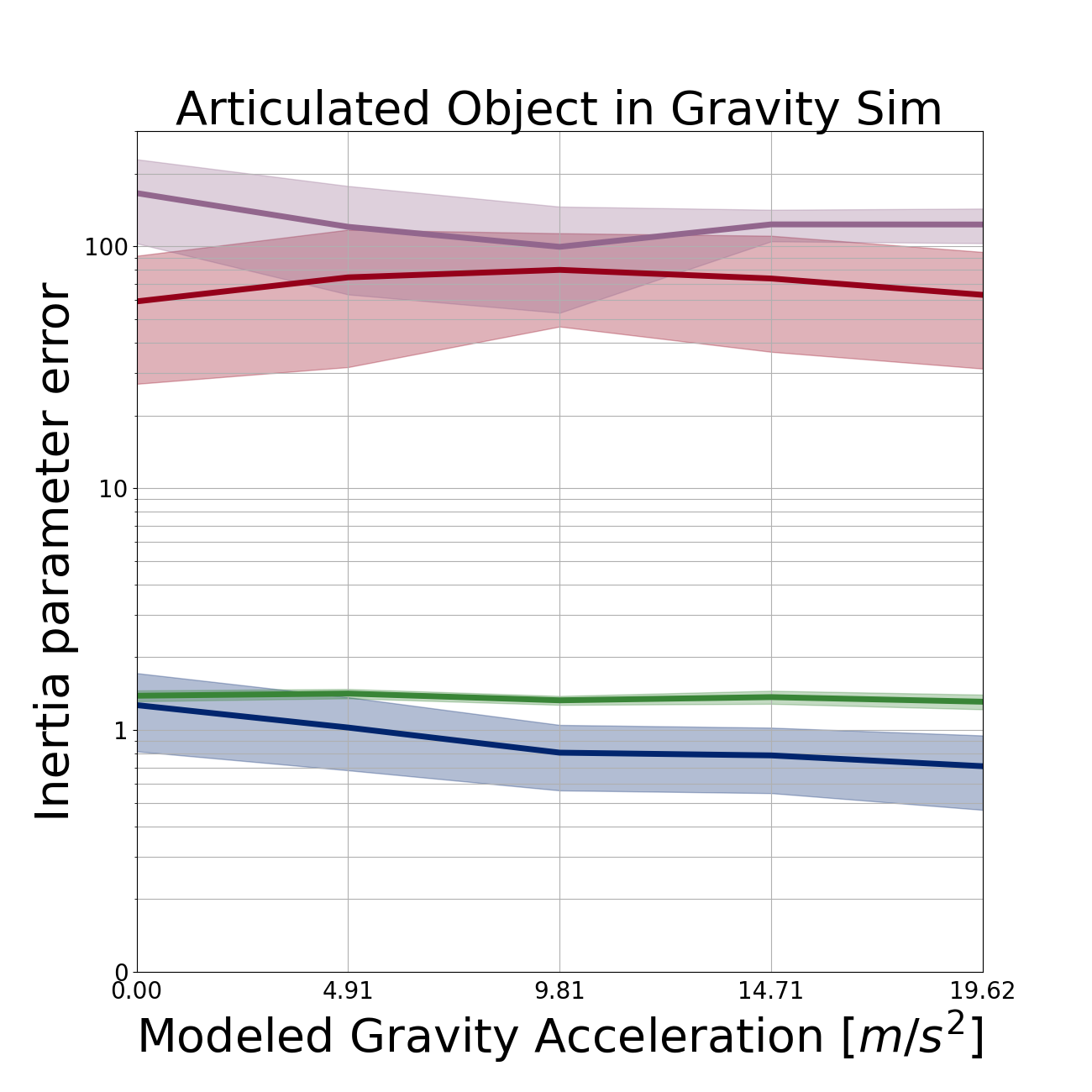}
    \vspace{-6mm}
    \caption{Friction (\textbf{Left}, \textbf{Left middle}) and inertia (\textbf{Right middle}, \textbf{Right}) parameter errors for the vortex and gravity simulated experiments.}
    \label{fig:parameter_errors}
    \vspace{-1.5em}
\end{figure}

\paragraph{Parameter Initializations.}  All experiments were run a minimum of 9 times each with a random parameter initialization using a process described in Appendix \ref{apx:initialization}.  Shaded regions in the results plots indicate 5\%/95\% normal t-score confidence intervals.

\paragraph{Comparisons and Evaluation Metrics.}
See Table \ref{table:methods} for the five approaches we tested.  Anitescu dynamics \citep{anitescu2006optimization} were selected as a reasonable differentiable simulation baseline, as it forms the basis of many widely-used, modern simulators, notably including MuJoCo \cite{todorov2012mujoco} and Drake \cite{drake}.  We present prediction errors for all approaches and parameter errors for the structured approaches.  Prediction errors are the average norm error of all bodies' position or orientation over the course of a trajectory.  Defining $\mathcal{V}$ as the set of points inside a body's geometry and $\mathcal{I} = [m, p_x, p_y, p_z, I_{xx}, I_{yy}, I_{zz}, I_{xy}, I_{xz}, I_{yz}]$ as the set of body inertial parameters, parameter errors for quantifying the geometry, friction, and inertial properties for a system with $n$ bodies are
\begin{subequations} \label{eqn:parameter_errors}
\vspace{-1em}
\begin{align}
    e_\text{volume} &= \frac{1}{n} \sum_{i=1}^n \frac{\text{Vol}\left( \left( \mathcal{V}_{i, \text{actual}} \setminus \mathcal{V}_{i, \text{learned}} \right) \cup \left( \mathcal{V}_{i, \text{learned}} \setminus \mathcal{V}_{i, \text{actual}} \right) \right)}{\text{Vol}\left( \mathcal{V}_{i, \text{actual}} \right)}, \label{eqn:volume_metric} \\
    e_\text{friction} &= \norm{\left[ \mu_1, \dots, \mu_n \right]_\text{learned} - \left[ \mu_1, \dots, \mu_n \right]_\text{actual}}, \\
    e_\text{inertia} &= \norm{\left[ s \cdot \mathcal{I}_1, \dots, s \cdot \mathcal{I}_n \right]_\text{learned} - \left[ s \cdot \mathcal{I}_1, \dots, s \cdot \mathcal{I}_n \right]_\text{actual}}. \label{eqn:inertia_metric}
\end{align}
\end{subequations}
The vector $s$ is akin to a \enquote{characteristic length} that is effectively normalized by the inertia of the true object.  See more details on the volume and inertia metrics in Appendices \ref{apx:volume_metric} and \ref{apx:inertia_metric}, respectively.


\section{Results}
\label{sec:results}


We test the methods across challenging datasets featuring collisions through contact-rich trajectories.  While the contact dynamics are prominent in all the example trajectories, we built the articulated system in particular for its continuous dynamics:  non-trivial due to state-dependent Coriolis and centrifugal effects and unmodeled joint friction, damping, or backlash.  

\paragraph{Observations.}
Both real experiments (articulated object and cube in left and left middle columns in Fig.\ \ref{fig:results}, respectively) show separation between CCN, DiffSim, and End-to-end methods, with CCN  matching or outperforming alternatives along all metrics, especially with more data.  On the cube dataset, CCN and CCN-R consistently converge to $<$10\% volume error while DiffSim struggles to improve even with more data.  The residual does not significantly help with real data but does in simulated examples, where CCN-R in the vortex scenario (right middle column in Fig.\ \ref{fig:results}) improves its positional trajectory error significantly, achieving consistently 5x better performance than other methods at the largest dataset size. On the same metric, DiffSim-R sees no improvement beyond DiffSim.  Fig.\ \ref{fig:parameter_errors} shows CCN and CCN-R nearly always outperforms DiffSim and DiffSim-R, except for the vortex scenario where all methods perform well. In the gravity scenario (right column in Fig.\ \ref{fig:results}), the residual helps CCN-R maintain good performance achieved at the correct gravitational acceleration, across all initial models.  In contrast, DiffSim-R outperforms DiffSim at every initial gravitational acceleration model.  Since this gravity scenario swept over different modeled gravitational accelerations, End-to-end is unaffected since its representation is unstructured, and its consistent performance over the x-axis of these plots are included for reference against CCN and DiffSim.  Parameter errors in Fig.\ \ref{fig:parameter_errors} indicate DiffSim and DiffSim-R struggled to capture both the friction and inertial terms in the gravity scenario to levels attained by CCN and CCN-R.

\paragraph{Implications.}  The residual's lack of effect on real data is in alignment with prior works that found the rigid body model already performs well in contact-rich scenarios with simple systems \citep{acosta2022validating}.  The residual shows the most merit in the more extreme simulation examples, though its effect isn't realized until larger dataset sizes -- unsurprising for a DNN.  The residual helps DiffSim to a much lesser extent because our method better separates continuous and contact dynamics and allows the residual to identify the smooth nature of the unmodeled vortex dynamics.  Relatedly, there is better performance for DiffSim and DiffSim-R at overestimated gravitational accelerations rather than underestimated, where contact is less often predicted.  Without contact, prediction losses experience a lack of informative parameter gradients, in which case DiffSim-R outperforms DiffSim.


\section{Conclusion}
\label{sec:conclusion}

We demonstrate with real experiments that our violation implicit loss trains models that outperform prediction loss-based structured and unstructured models.  Our approach leverages the structure of contact versus continuous dynamics to learn both simultaneously, with physically meaningful parameters driving separate contact and continuous dynamics with a DNN residual to augment.

\section{Limitations and Future Work.}
\label{sec:limitations}

The articulated object is the most challenging system presented herein for dynamics learning.  While our methods outperformed alternatives in all other metrics, there is still a significant gap between ground truth and our models’ trajectory predictions, and the rotational error showed lackluster performance from all methods.  Further closing this gap remains for future exploration, and we are open-sourcing our articulated object dataset for the community to contribute their own methods.  The scalability of the method in this paper has not yet been demonstrated on large-scale systems (e.g. a robotic arm) or in multi-object settings. It remains to be seen whether the advantages demonstrated here will extend as scope increases.  While we tested one version of differentiable simulation using \citet{anitescu2006optimization} dynamics, future studies will compare alternatives against each other and our violation implicit loss in performing system identification.  Our approach encouraged the residual to fill gaps in continuous dynamics while relying on a rigid body contact dynamics model to handle contact.  Other works have demonstrated improved prediction capability by learning the contact model \citep{allen2023graph}, though integrating this with system identification remains future work. While other works have learned articulated structures from scratch \citep{sturm2011probabilistic, ma2023sim2real, heppert2023carto}, we assumed access to kinematic structure in this paper, leaving joint kinematics/dynamics learning for future studies.  Lastly, we relied on AprilTags to estimate poses, which are more challenging to obtain via perception \citep{wen2021bundletrack, chen2023texpose, liu2022gen6d}.



\clearpage
\acknowledgments{We thank our anonymous reviewers, who provided thorough and fair feedback.  This work was supported by a National Defense Science and Engineering Graduate Fellowship, an NSF Graduate Research Fellowship under Grant No. DGE-1845298, and an NSF CAREER Award under Grant No. FRR-2238480.}


\bibliography{refs}  

\clearpage
\appendix

\section{Learning Details}
\label{apx:learning_details}

{\renewcommand{\arraystretch}{1.1}
\begin{table}[h]
    \centering
    \begin{tabular}{ | c | c | c | c | c | }
        \hline
        \textbf{Hyperparameter} & \textbf{CCN, real} & \textbf{CCN, sim} & \textbf{DiffSim, real} & \textbf{DiffSim, sim} \\ \hline \hline
        $w_\text{comp}$ & 0.001 & 0.001 & N/A & N/A \\
        $w_\text{diss}$ & 0.1 & 0.1 & N/A & N/A \\
        $w_\text{pen}$ & 100 & 100 & N/A & N/A \\
        $w_\text{res}$ & 1 & 0.001 & 1000 & 1 \\
        $w_\text{res, w}$ & 0.1 & 0 & 1 & 0 \\ \hline
    \end{tabular}
    \caption{Tuned hyperparameters.  Rows for residual norm ($w_\text{res}$) and weight ($w_\text{res, w}$) regularization only apply for -R variations.  Real versus simulated experiments performed best with different residual regularization weights since the simulations featured larger model-to-actual dynamics gaps.}
    \label{table:hyperparameters}
\end{table}
}

\subsection{Model-Based Parameter Learning}
\label{apx:model_tuning}
For our CCN and CCN-R method, we performed a hyperparameter search to determine the most effective set of weights for balancing the loss terms in \eqref{eqn:h_form}.  See Table \ref{table:hyperparameters} for these sets of weights.

\subsection{Residual Network Architecture and Regularization}
\label{apx:residual}
The residual network featured in both CCN-R and DiffSim-R has the same architecture.  The first layer takes in the full state of the system and converts the quaternion orientation representation into a 9-vector of the elements of the corresponding rotation matrix, letting the remaining state positions and velocities pass through to layer 2. Beyond the first layer, the network is a fully-connected multi-layer perceptron (MLP) with two hidden layers of size 128.  The last layer outputs values in the acceleration space of the system.  All activations are ReLU.

We regularized the residual via both output norm regularization and weight regularization, with associated weight hyperparameters $w_\text{res}$ and $w_\text{res, w}$, respectively.  See Table \ref{table:hyperparameters} for the optimal values.  Since the simulation examples were specifically designed to test the capabilities of the residual network, we found the optimal weights for the residual terms were much lower for simulated examples than for the real data.  We also note that the optimal residual weights were much higher for DiffSim than for CCN.  This is a direct result from the DiffSim residual's attempts to explain some of the contact dynamics, whose accelerations are orders of magnitude larger than the continuous accelerations.  Our CCN method avoids this by better containing its residual in the continuous domain, and thus could use lower residual regularization weights.

\subsection{End-to-End Network Architecture}
\label{apx:end_to_end}
The best performing network for the End-to-end baseline is an MLP with 4 hidden layers each of size 256 with Tanh activation.  Its input is the full state of the system, and its output is the next velocity.  The next configuration is obtained from predicted next velocity with an Euler step \eqref{eqn:config_pred}.

\subsection{Parameter Initializations}
\label{apx:initialization}
All learned parameters are randomly initialized within pre-specified ranges. Geometric parameters are initialized between 0.5 and 1.5 times their true lengths, and friction coefficients between 0.5 and 1.5 times their approximate true values. Inertial parameters are initialized to a set of physically feasible values via the following procedure for each link in the body:
\begin{enumerate}
    \item A virtual link is sized via a random set of three length scales $l_x, l_y, l_z$, chosen between 0.5 and 1.5 times the link’s true dimensions.
    \item The center of mass of the link is initialized to be somewhere within the inner half of this virtual link’s geometry.
    \item A random mass $m_\text{rand}$ is selected from the range between $0.5 m_\text{urdf}$ and $1.5 m_\text{urdf}$.
    \item Principal axis moments of inertia are computed using the assumption of uniform density throughout the randomly-sized virtual link, via e.g. for $I_{xx}$:
    \begin{align} I_{xx, \text{principal axis}} &= \frac{m_\text{rand}}{12} \left( l_y^2 + l_z^2 \right).\end{align}
    \item Rotate the inertia matrix from its principal axis definition by a random rotation in SO(3). The link’s initialized moments and products of inertia are derived from this rotated version.
\end{enumerate}

\section{Evaluation Metric Details}
\label{apx:eval_metrics}

\subsection{Volume Evaluation Metric}
\label{apx:volume_metric}
The volume error metric defined in Equation \eqref{eqn:volume_metric} is computed as the fraction of volume that the learned geometry incorrectly included or incorrectly excluded. To compute this, we used the identity
\begin{align}
    \mathrm{Vol}(A \setminus B) &= \mathrm{Vol}(A) - \mathrm{Vol}(A  \cap B).
\end{align}
The numerator of \eqref{eqn:volume_metric} can therefore be computed as
\begin{align}
    \mathrm{Vol}(\mathcal V_{i,\text{actual}}) + \mathrm{Vol}(\mathcal V_{i,\text{learned}}) - 2\mathrm{Vol}(\mathcal V_{i,\text{actual}}  \cap \mathcal V_{i,\text{learned}}).
\end{align}
$\mathcal V_{i,\text{learned}}$, $\mathcal V_{i,\text{actual}}$, and their intersection are all convex hulls of a finite number of vertices. Therefore, the interaction operation as well as volume calculation can be conducted with a standard convex hull or halfspace intersection library, such as \texttt{qhull}.

\subsection{Inertia Evaluation Metric}
\label{apx:inertia_metric}
A body's set of inertial parameters is $\mathcal{I} = [m, p_x, p_y, p_z, I_{xx}, I_{yy}, I_{zz}, I_{xy}, I_{xz}, I_{yz}]$.  Since true inertia parameter vectors feature values at wildly different scales, the vector $s$ is selected to normalize $\mathcal{I}$ to more equally evaluate all inertial parameter errors.  For example, the true inertial parameters for the simulated asymmetric object used in the vortex example are
\begin{align}
    \mathcal{I}_\text{asym} &= \left[ 0.25, 0, 0, 0, 0.00081, 0.00081, 0.00081, 0, 0, 0 \right].
\end{align}
Choosing 3.5cm as a reasonable center of mass location distance, the associated $s_\text{asym}$ normalizer is
\begin{align}
    s_\text{asym} &= \left[ \frac{1}{0.25}, \frac{1}{0.035}, \frac{1}{0.035}, \frac{1}{0.035}, \frac{1}{0.00081}, \frac{1}{0.00081}, \frac{1}{0.00081}, \frac{1}{0.00081}, \frac{1}{0.00081}, \frac{1}{0.00081} \right].
\end{align}

\end{document}